\documentclass{article}

\usepackage[preprint]{neurips_2026}

\usepackage[utf8]{inputenc}
\usepackage[T1]{fontenc}
\usepackage{hyperref}
\usepackage{url}
\usepackage{booktabs}
\usepackage{amsfonts}
\usepackage{amsmath}
\usepackage{nicefrac}
\usepackage{microtype}
\usepackage{xcolor}
\usepackage[dvipsnames]{xcolor}
\usepackage[normalem]{ulem}
\usepackage{graphicx}
\usepackage{tikz}
\usetikzlibrary{shapes.geometric, arrows, fit, positioning, backgrounds}

\tikzstyle{startstop} = [rectangle, rounded corners, minimum width=1.5cm, minimum height=1cm,text centered, draw=black, very thick, fill=red!30]
\tikzstyle{greybox}   = [rectangle, rounded corners, minimum width=1.5cm, minimum height=1cm,text centered, draw=black, very thick, fill=gray!30]
\tikzstyle{every fit} = [inner sep=0.2cm]
\tikzstyle{io}        = [trapezium, trapezium left angle=70, trapezium right angle=110, minimum width=1.5cm, minimum height=1cm, text centered, draw=black, fill=blue!30]
\tikzstyle{process}   = [rectangle, rounded corners, minimum width=1.5cm, minimum height=1cm, text centered, draw=black, very thick, fill=orange!30]
\tikzstyle{record}    = [rectangle, rounded corners, minimum width=1.5cm, minimum height=1cm, inner sep=0.2cm, text centered, draw=black, very thick, fill=blue!30]
\tikzstyle{process2}  = [rectangle, rounded corners, minimum width=1.5cm, minimum height=1cm, text centered, draw=black, very thick, fill=green!30]
\tikzstyle{loss}      = [rectangle, rounded corners, minimum width=1.5cm, minimum height=1cm, text centered, draw=black, very thick, fill=magenta!30]
\tikzstyle{decision}  = [diamond, minimum width=1.5cm, minimum height=1cm, text centered, draw=black, fill=green!30]
\tikzstyle{arrow}     = [very thick,->,>=stealth]
\tikzstyle{varholder} = [minimum width=1.5cm, minimum height=0.75cm, draw=none, fill=none, align=center]

\title{Artificial Empathy: Towards a Framework for Unsupervised Agency Detection and Policy Reconstruction}

\author{%
  Peter Kuhn \\
  Human Inductive Bias Project \\
  Cambridge, UK \\
  \texttt{peterkuhn92@gmail.com} \\
  \And
  Chris Pang \\
  Human Inductive Bias Project \\
  Cambridge, UK \\
  \texttt{mutatismutandisetplusultra@gmail.com} \\
  \And
  Sonakshi Chauhan \\
  Human Inductive Bias Project \\
  Cambridge, UK \\
  \texttt{sonakshichauhan1402@gmail.com} \\
}

\begin{document}

\maketitle

\begin{abstract}
We study how an AI system can identify and model other agents in its environment from observation alone, which is a capability necessary for cooperative behaviour in the real world. This problem is less constrained than inverse reinforcement learning and remains largely unexplored. We propose a framework that uses a reinforcement learning agent, trained on an independent task as a prior about agentic dynamics, to perform agency detection and policy reconstruction.
\end{abstract}

\section{Introduction}

We define agency detection as the task of distinguishing between agentic behaviour and non-agentic activity in an environment. We define policy reconstruction as the task of inferring the policy of an agent from behaviour alone. The latter task, in isolation, is typically tackled in inverse reinforcement learning (IRL), but there are few prior attempts \citep[for an exception, see][]{zarncke2025foundations} to solve both tasks in conjunction. If we could implement agency detection and policy reconstruction in an unsupervised way, this would be a step towards solving the AI alignment problem, the problem of aligning the goals of AI systems with the goals of their human creators \citep{bostrom2014superintelligence}. An agent equipped with such a capacity could then develop and act on a joint policy that is a synthesis of its own policy and the policy of the agents in its environment. Such a system would, so to speak, have the propensity to care for other agents. We develop \textit{artificial empathy} (AE) as a framework for solving this task.

Implementing artificial empathy is hard: \textit{a priori}, every random physical event could be the result of agentic activity. Did the apple \textit{want} to fall off the table? To decide this, a vast array of prior knowledge about the nature of apples and the nature of agency is required. Inspired by psychological models of empathy, we suggest that agency detection can work by finding systems that one can model using \textit{one's own} agential dynamics: As the AI system that we want to align is itself a form of agent, it can use its dynamics as a prior about what agency looks like. Thus, a system implementing artificial empathy jointly solves the agency-detection and policy-reconstruction tasks by determining whether parts of the environment can be predicted from variations in its own dynamics, and hence be considered agents. In this paper, we take the first steps towards implementing artificial empathy: We propose an architecture that uses a trained model-based reinforcement learning (RL) agent as a prior to perform agency-detection and policy reconstruction. We demonstrate that this system can be used to differentiate between simple agentic activity and random dynamics.

\section{Prior Work}

\subsection{Psychological Inspirations and Self-Other Overlap}
We take inspiration for our approach from simulation theory in cognitive science, which suggests that agents understand and predict others' behaviour by using their own cognition and decision-making rather than an independent theory of mind. \citet{https://doi.org/10.1111/j.1468-0017.1986.tb00324.x} introduced this in the context of folk psychology, claiming that humans often ``put themselves in the other's place'' to infer and explain the behaviour of others. The plausibility of simulation-based mechanisms for social understanding is supported by neuroscientific findings, such as mirror neuron responses, in which a population of neurons activates both during execution and observation of an activity \citep{gallese1998mirror, goldman2006simulating}.

We are also inspired by alignment approaches that reduce self-other overlap in LLM-based agents by reducing the distance between the agent's self-representation and representations of other agents or humans \citep{carauleanu2024safehonestaiagents}. However, even though inducing such overlap can produce behaviour that appears to be aligned, the method is primarily intuition-driven and is predicated on accurately identifying self- and other-representations in activation space. By establishing artificial empathy as a first-principles framework, we hope to overcome these limitations.

\subsection{Inverse Reinforcement Learning}

Inverse RL (IRL) is a well-established approach that aligns with our notion of artificial empathy. It involves inferring an agent's reward function from observed behaviour, typically given as labelled observation--action pairs \citep{irl}. Extensions using maximum-entropy priors have yielded promising results despite the under-constrained nature of the inference problem \citep{entropyirl}. To reduce goal misgeneralization, IRL has also been proposed as a step towards AI alignment \citep{russell2019human}. However, IRL relies on preprocessed observation-action trajectories, limiting its applicability to data-rich settings where agent behaviour is clearly identifiable. Thus, IRL and its variants cannot be complete alignment strategies in themselves. 

\section{Method}

\subsection{Overview and Motivation}

Our goal is to infer the policy of an agent from observing the environment it inhabits. In our current setup, we will assume full observability of the environment, represented by observations $(o_0, o_1, \ ldots)$, for ease of implementation. Nothing deep hinges on this. We assume that the environment contains an agent, the ``other", following some policy $\pi_O$. The task of the AE system is to infer $\pi_O$ from the stream of observations alone. We differentiate between base agents and the AE system, where the former are RL systems trained to play specific games, and the latter aims to solve the AE task. In our setup, the AE task is purely epistemic, i.e. we want it to infer what is an agent and what is it trying to do. The broader aim of AE is the creation of a \emph{joint policy} $\pi_\Omega$ that joins the policy of the agent itself (``self") $\pi_S$ and the policy of the ``other" $\pi_O$ using some aggregation function $f$, i.e. $\pi_\Omega=f(\pi_S,\pi_O)$. We use a deep Q-learning framework to perform RL and to represent policies in terms of Q-functions $Q(s,a)$ for some state $s$ and action $a$ \citep{sutton2018reinforcement}. As these Q-functions will, in future work, implement complex joint policies learned from other agents, the relevant states $s$ should refer to hidden states of the environment (in the sense of a partially observable Markov decision process, see \citet{puterman2014markovdecisionprocesses}) rather than simple observations. If you want to infer what some agent cares about, you first need to construct a convergent representation of the joint environment. This is why we use a model-based RL setup \citep{moerland2022modelbasedreinforcementlearningsurvey}. We will first describe the base agent that will serve both as the target ``other" in our experiments, as well as an agency prior. Note that we provide a diagram in Appendix~\ref {sec:appendix}.

\subsection{Base Agent Architecture}

Our base architecture is a deep Q-learner \citep{mnih2015humanlevelcontrol} that we have augmented with a trainable world model. The world model is inspired by a variational autoencoder \citep{kingma2022autoencodingvariationalbayes, hafner2025trainingagentsinsidescalable} that infers a probabilistic representation of world states. We use neural-network-based models that represent probability distributions as Gaussians parametrised by means and logarithmic variances. The architecture consists of a convolutional network encoder $p_\theta(s_t|o_t)$ parametrized by $\theta$ that infers world states $s_t$ from observations $o_t$, a dense transition model $q_\phi(s_{t+1}|s_t,a_t)$ parametrized by $\phi$ that predicts the next world state $s_{t+1}$ depending on the action taken $a_t$, and a convolutional, non-probabilistic decoder $d$ parametrized by $\omega$ that reconstructs observations based on world states $\hat{o}_{t+1}=d_\omega(\hat{s}_{t+1})$. Here $\hat{s}_{t+1}$ is sampled from the transition model. Finally, the base agent contains a Q-network that predicts Q-values for world states $Q_\rho(s_t)$, parametrized by $\rho$. Note that $Q(s_t,a_t)$, necessary for the sort of Q-learning we want to do, can be recovered by using the transition model to calculate Q-values over states and actions. In later experiments, we train the base agents using a variant of the experience-replay algorithm described in \cite{mnih2015humanlevelcontrol} where at every time step $t$ the agent sees two observations, the action it has taken, and a reward $(o_t,a_t,o_{t+1},r_{t+1})$ and integrates the models by combining three kinds of losses. The reconstruction loss $\mathcal{L}_{recon}$ is defined as the mean squared error between the true next observation $o_{t+1}$ and its reconstruction $\hat{o}_{t+1}$, where $\hat{o}_{t+1}$ is sampled from the model; the transition loss $\mathcal{L}_{trans}$ is the Kullback--Leibler divergence between the modeled distribution $p_\theta(s_{t+1}\mid o_{t+1})$ and the approximate posterior $q_\phi(s_{t+1}\mid s_t,a_t)$ over the next latent state $s_{t+1}$ given the current latent state $s_t$ and action $a_t$; and the Q-loss $\mathcal{L}_Q$ is the mean squared error between the target value $y_i$ and the predicted value $Q_i$. With $y_i = r_i + \gamma \max_{a'} Q(s'_i, a')$ and $Q_i = Q(s_i, a_i)$. $i$ indexes transitions in a mini-batch. This results in a combined loss for the base agent:

\begin{equation}
    \mathcal{L}_{base}=\alpha \mathcal{L}_{recon}+\beta \mathcal{L}_{trans}+\delta \mathcal{L}_{Q}
\end{equation}

We chose the hyperparameters $\alpha$, $\beta$, $\delta$ and $\gamma$ to be $10^{-4}$, $10^{-2}$, $1$ and $0.98$ respectively.

\subsection{AE Architecture}

The AE system uses a trained base agent as an agency-prior. Thus, it is structurally similar to the base agent and also consists of a similar encoder, transition and Q-network combination. Learning proceeds by observing a base agent trained to perform some task. We will call the first base agent the prior agent, and the game it learned the prior game. The target base agent we will call the expert. The environment in which the expert acts generates the stream of observations $(o_0, o_1, ...)$. These are reordered by the AE system and used as training data. The task is to infer the expert's policy by reconstructing its encoder, transition model, and Q-network. We initialise an AE system by loading a prior, consisting of the parameters $(\bar{\theta}, \bar{\phi}, \bar{\rho})$ corresponding to the weights of the prior agent. We will assume a Gaussian prior probability in weight-space for every network with a mean around the prior weights. As is known from a Bayesian analysis of weight decay, a prior Gaussian in weight-space can be enforced by the use of mean squared error \citep{mckay}. We will express the prior constraints on our model as set of penalty terms $\mathcal{R}_p$, $\mathcal{R}_q$ and $\mathcal{R}_Q$ for encoder, transition model and Q-network respectively, which are the mean square error between the weights of the prior agent's parameters and the parameters of the AE system.

Conceptually, we want to predict $s_{t+1}$ from $s_t$ (i.e. the dynamics of the world) by assuming that $s_{t+1}$ was the result of an agent observing the environment and acting on it. To do this, we will use networks constrained by the agency prior, which we will distinguish from those of the base agent with a dash. We encode the future observations as before as $p_{\hat{\theta}}(s_{t+1}|o_{t+1})$. We then make a guess about how the agent putatively acting on the environment perceives the world by encoding the current observation as $p'(s_t|o_t)$. We then make predictions about the results the agent we are modelling might expect to find for each action $a_i$ by computing $q'(s_{t+1}|s_t,a_i)$. We can then make a prediction about how much the modelled agent values different states of the world by computing $Q'(s_t,a_t)$. By taking a softmax, we turn this into probabilities for discrete actions taken by the putative agent $P(a_i)$.

Now we have all the ingredients we need to predict the next state of the world, under the assumption that these are generated by an agent (denoted by $A$) following some policy. To do this, we simply marginalise over all $N$ different actions to obtain a probability distribution over next world states. We use $p^*$ to differentiate it from the results of an encoding:

\begin{equation}
    p^*(s_{t+1}|s_t, A)=\sum_{i=0}^{N}P(a_i)q'(s_{t+1}|s_t,a_i)
\end{equation}

Thus, we can naturally use the Kullback-Leibler divergence between the probability distribution of encoded world-states and the world-states we would expect on the assumptions that there is an agent acting in the environment, parametrized in the way described. Thus the main loss of our AE system is:

\begin{equation}
\mathcal{L}_{KL}=D_{KL}[p(s_{t+1}|o_{t+1})||p^*(s_{t+1}|s_t, A)]
\end{equation} 

Finally, to get the full loss of the artificial empathy system, we have to ensure that our system actually \textit{commits} to some predicted action, rather than just returning a smooth probability distribution and predicting the world dynamics from there. We do this by putting an entropy penalty on $P(a_i)$, i.e. $\mathcal{R}_H =H(P(a_i))$. The combined loss for the AE-system is thus:

\begin{equation}
\mathcal{L}_{AE}=\mathcal{L}_{KL} + \sigma\mathcal{R}_H + \kappa\mathcal{R}_p + \psi\mathcal{R}_q +\upsilon\mathcal{R}_Q 
\end{equation} 

In words: We want to infer the world dynamics under the assumption that an agent is present, parametrised similarly to the prior agent, and that takes specific actions. We choose the hyperparameters $\sigma$, $\kappa$, $\psi$ and $\upsilon$ to be $10$, $100$, $1000$ and $1000$ respectively.

\section{A Simple Experiment}

We validate the feasibility of the proposed architecture as follows. We train two base agents on tasks $A$ and $B$. We use the agents trained on $A$ as an expert and $B$ as a prior to be loaded into the AE system. After training on fully observable trajectories, the AE system is then evaluated on its ability to imitate the expert by selecting actions via its learned Q-values. Performance in the `imitation game' is measured by cumulative reward on task $A$. Experiments on the imitation game are conducted in a 15$\times$15 gridworld with two tasks. In the food-game, the agent must collect three food items, each yielding a reward of 1, while every step incurs a reward of -0.01. In the edge-game, the agent must reach the lower-left edge to obtain a reward of 1, again receiving a reward of -0.01 for every step taken. Base agents are trained for $1200$ epochs for stability and use $\epsilon$-decay with $10^5$ decay steps starting at $1$.

\begin{figure}[t]
  \begin{center}
    \centerline{\includegraphics[width=0.5\columnwidth]{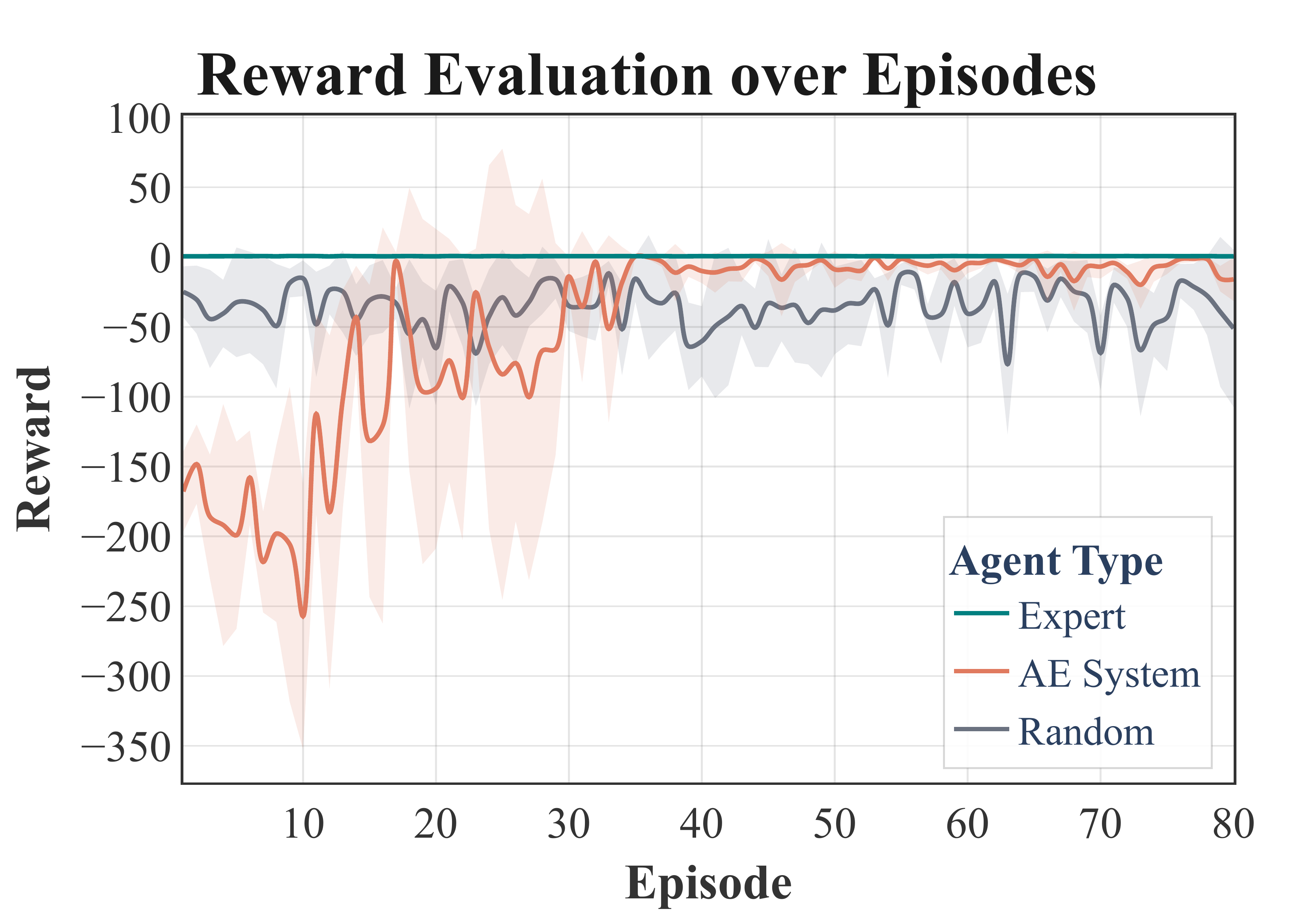}}
    \caption{
      Aggregate of 5 runs of the `imitation game'. Shaded regions are standard deviations. After 35 episodes the performance of the AE agent is far better than a random walk and approaches that of the base agent.
    }
    \label{food2edge}
  \end{center}
\end{figure}

The observations generated by these games are 2x15x15 tensors, where the first channel encodes food item positions (or nothing for the edge-game) and the agent position. We use the food-game as the prior game $B$ and the edge-game as task $A$. To make the task of the AE-system less simple, we confound its observation stream by adding a decoy `agent' that merely performs a random walk. If the AE agent can nonetheless learn to imitate the edge-game agent, then it has implicitly solved the task of agency detection. As shown in Figure \ref{food2edge}, this is in fact what we observe. Note that this result is far from trivial: To achieve it, the agent has to pick up on the fact that the random dynamics of the decoy are ill-suited for prediction via the prior implicit in the AE-system. It has to utilise implicit prior knowledge about how different actions lead to different world-states, but it has to shift the implicit value of world-states to accommodate the observed behavioural profile. While these are early results, we consider them quite promising.

\section{Limitations}

Our experiments are conducted in a small, fully observable gridworld with a single confounding decoy agent, so the current results demonstrate feasibility rather than robustness or scale. The AE system currently relies on a hand-selected, structurally similar prior agent; how performance degrades as the prior and the target policy diverge in architecture or task structure remains untested. We also did not yet evaluate settings with multiple genuine agents, partial observability, or non-stationary policies, all of which are more representative of real-world alignment settings. Finally, because the AE loss combines several weighted terms ($\mathcal{L}_{KL}$, $\mathcal{R}_H$, $\mathcal{R}_p$, $\mathcal{R}_q$, $\mathcal{R}_Q$), the sensitivity of results to these hyperparameters has not been systematically characterised.

\section{Conclusion and Future Work}

We introduced artificial empathy, a framework that jointly performs agency detection and policy reconstruction by using an RL agent's own dynamics as a prior over agentic behaviour. In a simple gridworld imitation task with a random-walk decoy, an AE system initialised from an unrelated prior task learned to imitate an expert agent's policy, suggesting that self-referential priors can support unsupervised agency detection. Future work would extend the framework to conditions of partial observability, scaling to environments with multiple simultaneous agents, relaxing the structural similarity between prior and target agents, using agency priors trained on multiple games, and moving from the purely epistemic AE task described here towards the construction of an explicit joint policy $\pi_\Omega$ that combines self and other.

\bibliography{iclr2026_conference}
\bibliographystyle{plainnat}

\appendix
\section{Artificial Empathy System Diagram}
\label{sec:appendix}

Diagram displaying the structure of the base agent and the AE system. Overlapping layers of boxes represent computation for many possible actions: the system computes one predicted future state for each action and assigns each predicted state a Q-value separately. The state with the maximum Q-value is then selected via argmax in the base architecture, and the states are marginalised via a softmax in the AE system.

\scalebox{0.75}{
\begin{tikzpicture}[node distance=1.2cm]

\node (o_t) [varholder] at (0,0) {$o_t$};

\node (enc) [startstop, below of=o_t, yshift=-0.5cm] {Encoder};
\draw[arrow] (o_t) -- (enc);

\node (pro3) [process, below of=enc, yshift=-0.2cm, xshift=0.2cm, opacity=0.3] {Transition};
\node (pro2) [process, below of=enc, yshift=-0.35cm, xshift=0.1cm, opacity=0.5] {Transition};
\node (pro1) [process, below of=enc, yshift=-0.5cm] {Transition};

\draw[arrow] (enc) -- (pro1);

\node (q3) [process2, below of=pro1, yshift=-0.2cm, xshift=0.2cm, opacity=0.3] {Q-value};
\node (q2) [process2, below of=pro1, yshift=-0.35cm, xshift=0.1cm, opacity=0.5] {Q-value};
\node (q1) [process2, below of=pro1, yshift=-0.5cm] {Q-value};

\node (dec) [startstop, right of=q1, xshift=1cm] {Decoder};

\node (a_t) [varholder, below of=q1, yshift=-0.5cm] {$a_t$};

\node (hat_o_t+1) [varholder, below of=dec, yshift=-0.5cm] {$\hat{o}_{t+1}$};

\node (qloss) [loss, below of=a_t, yshift=-0.2cm] {Q Loss};

\node (reward) [varholder, above left=0.4cm and 0.0cm of qloss] {$r_{t+1}$};
\draw[arrow] (reward) -- (qloss);

\node (ploss) [loss, below of=hat_o_t+1, yshift=-0.2cm] {MSE Loss};

\node (o_t+1) [varholder, right of=o_t, xshift=3.5cm] {$o_{t+1}$};
\node (enc2) [startstop, below of=o_t+1, yshift=-0.5cm] {Encoder};

\node (kldloss) [loss, right of=ploss, xshift=1.4cm] {$D_{KL}$ Loss};

\draw[arrow] (o_t+1) -- (enc2);

\draw[arrow] (pro1) to[bend left=35]
    node[pos=0.78, fill=gray!30, inner sep=1pt, yshift=1cm, xshift=-0.5cm]
        {$p(s_{t+1}\mid s_t)$}
    (kldloss);

\draw[arrow] (enc2) --
    node[pos=0.78, fill=white, inner sep=1pt, yshift=-2pt]
        {$p(s_{t+1}\mid o_{t+1})$}
    (kldloss);

\draw[arrow] (pro1) -- (q1);
\draw[arrow] (pro1) edge [bend left=45] (dec);
\draw[arrow] (q1) -- (a_t);
\draw[arrow] (a_t) -- (qloss);
\draw[arrow] (dec) -- (hat_o_t+1);
\draw[arrow] (hat_o_t+1) -- (ploss);

\node (o_t+1_mse) [varholder, above of=ploss, xshift=1.0cm, yshift=0.2cm] {$o_{t+1}$};
\draw[arrow] (o_t+1_mse) -- (ploss);

\begin{scope}[on background layer]
    \node (agentbox) [greybox, fit=(enc) (pro1) (pro2) (pro3) (q1) (q2) (q3) (dec) (enc2)] {};
    \node (agent) [varholder, above of=agentbox, align=center, yshift=1.5cm] {\small \emph{Base Agent}};
\end{scope}

\node (o_t_right) [varholder, right of=o_t+1, xshift=2.75cm] {$o_t$};

\node (ae_enc) [startstop, below of=o_t_right, yshift=-0.5cm] {Encoder};
\draw[arrow] (o_t_right) -- (ae_enc);

\node (ae_pro3) [process, below of=ae_enc, yshift=-0.2cm, xshift=0.2cm, opacity=0.3] {Transition};
\node (ae_pro2) [process, below of=ae_enc, yshift=-0.35cm, xshift=0.1cm, opacity=0.5] {Transition};
\node (ae_pro1) [process, below of=ae_enc, yshift=-0.5cm] {Transition};

\draw[arrow] (ae_enc) -- (ae_pro1);

\node (ae_q3) [process2, below of=ae_pro1, yshift=-0.2cm, xshift=0.2cm, opacity=0.3] {Q-value};
\node (ae_q2) [process2, below of=ae_pro1, yshift=-0.35cm, xshift=0.1cm, opacity=0.5] {Q-value};
\node (ae_q1) [process2, below of=ae_pro1, yshift=-0.5cm] {Q-value};

\node (ae_dec) [startstop, right of=ae_q1, xshift=1.2cm] {Marginalise};
\draw[arrow] (ae_q1) -> (ae_dec);

\node (a_t_right) [varholder, below of=ae_q1, yshift=-0.5cm] {$P(a_t)$};

\node (hat_o_t+1_right) [varholder, below of=ae_dec, yshift=-0.5cm] {$p(s_{t+1}|A)$};

\node (qloss_right) [loss, below of=a_t_right, yshift=-0.2cm] {Entropy Loss};

\node (ploss_right) [loss, below of=hat_o_t+1_right, yshift=-0.2cm] {$D_{KL}$ Loss};

\node (o_t+1_right) [varholder, right of=o_t_right, xshift=4cm] {$o_{t+1}$};
\node (enc2b) [startstop, below of=o_t+1_right, yshift=-0.5cm, opacity=0.5] {Encoder Prior};
\draw[arrow] (o_t+1_right) -- (enc2b);

\draw[arrow] ([xshift=0.2cm]enc2b) to[bend left=30]
    node[pos=0.78, fill=white, inner sep=1pt, yshift=-2pt]
        {$p(s_{t+1}\mid o_{t+1})$}
    (ploss_right);

\begin{scope}[on background layer]
    \node (ae_agentbox) [greybox, fit=(ae_enc) (ae_pro1) (ae_pro2) (ae_pro3) (ae_q1) (ae_q2) (ae_q3) (ae_dec) (enc2b)] {};
    \node (ae_agent) [varholder, above of=ae_agentbox, align=center, yshift=1.5cm] {\small \emph{AE System}};
\end{scope}

\draw[arrow] (ae_pro1) -- (ae_q1);
\draw[arrow] (ae_pro1) edge [bend left=45] (ae_dec);
\draw[arrow] (ae_q1) -- (a_t_right);
\draw[arrow] (ae_dec) -- (hat_o_t+1_right);
\draw[arrow] (a_t_right) -- (qloss_right);
\draw[arrow] (hat_o_t+1_right) -- (ploss_right);

\draw[arrow] (agentbox) edge[dotted] (ae_agentbox) node[xshift=4.25cm, yshift=0.25cm] {\emph{Prior  }};

\end{tikzpicture}
}

\end{document}